\documentclass[letterpaper, 10 pt, conference]{ieeeconf}

\IEEEoverridecommandlockouts                              % This command is only needed if 
\makeatletter
\let\NAT@parse\undefined
\makeatother

\usepackage{amssymb,amsmath}

\usepackage[sort&compress,numbers,sectionbib]{natbib}
\usepackage{cite} % Compacts [1]--[4] citations and adds spaces
\usepackage{booktabs}
\usepackage{subcaption}
\usepackage{graphicx}

\usepackage{verbatimbox}
\usepackage{multirow}
\usepackage{balance}
\usepackage{wrapfig}

\newcommand{\figref}[1]{Fig.~\ref{#1}}
\usepackage[hyphens]{url}

\usepackage{hyperref}
\hypersetup{bookmarksopen,bookmarksnumbered,
pdfpagemode=UseOutlines,
colorlinks=true,
linkcolor=teal,
anchorcolor=teal,
citecolor=teal,
filecolor=teal,
menucolor=teal,
urlcolor=teal,
breaklinks=true
}
\usepackage[dvipsnames]{xcolor}
\definecolor{myred}{RGB}{215,48,39}
\definecolor{myblue}{RGB}{69,117,180}
\definecolor{myorange}{RGB}{252,141,89}
\definecolor{mylightblue}{RGB}{145,191,219}

\definecolor{MYlightblue}{RGB}{217,95,2} 
\definecolor{MYdarkblue}{RGB}{117,112,179} 
\definecolor{MYgreen}{RGB}{27,158,119}

\usepackage{color,soul} % Added by KB, for highlighting
\definecolor{krishna}{rgb}{.91,.65,.60} % Added by KB, for hl color
\definecolor{krishna-text}{rgb}{.59,.24,.19} % Added by KB, for note color
\usepackage{svg}

\newcommand{\xxnote}[3]{}
\ifx\hidenotes\undefined
  \usepackage{color}
  \renewcommand{\xxnote}[3]{\color{#2}{#1: #3}}
\fi

\usepackage[normalem]{ulem}
\useunder{\uline}{\ul}{}
\usepackage{todonotes}

\let\labelindent\relax

\usepackage{enumitem}
\usepackage{bm}


\usepackage{microtype}

{\title{
\LARGE \bf PuSHR: A Multirobot System for Nonprehensile Rearrangement}}

\author{Sidharth Talia$^{*}$, Arnav Thareja$^{*}$, Christoforos Mavrogiannis, Matt Schmittle, Siddhartha S. Srinivasa% <-this % stops a space
\thanks{$^*$Denotes equal contribution.}
\thanks{This work was (partially) funded by the Honda Research Institute USA, the National Science Foundation NRI (\#2132848) and CHS (\#2007011), DARPA RACER (\#HR0011-21-C-0171), the Office of Naval Research (\#N00014-17-1-2617-P00004 and \#2022-016-01 UW), and Amazon.}
\thanks{Authors are with the Paul G. Allen School of Computer Science \& Engineering, University of Washington, Seattle, WA, USA. Email: {\tt\small \{sidtalia, athareja, cmavro, schmttle, siddh\}@cs.washington.edu}.}%
}

\begin{document}

\maketitle
\thispagestyle{empty}
\pagestyle{empty}

% We focus on the problem of rearranging a cluttered environment via push-based manipulation carried out by a team of nonholonomically constrained mobile robots.
%The constraints from robot kinematics, workspace bounds, collision avoidance, and pushing stability demand highly coordinated robot motion that is challenging to plan and execute. 

% While there is extensive literature on multi-robot manipulation, prior work often assumes holonomic pushers, specialized grippers, and collaborative transport settings. 

\begin{abstract}
We focus on the problem of rearranging a set of objects with a team of car-like robot pushers built using off-the-shelf components. Maintaining control of pushed objects while avoiding collisions in a tight space demands highly coordinated motion that is challenging to execute on constrained hardware. Centralized replanning approaches become intractable even for small-sized problems whereas decentralized approaches often get stuck in deadlocks. Our key insight is that by carefully assigning pushing tasks to robots, we could reduce the complexity of the rearrangement task, enabling robust performance via scalable decentralized control. Based on this insight, we built PuSHR, a system that optimally assigns pushing tasks and trajectories to robots offline, and performs trajectory tracking via decentralized control online. Through an ablation study in simulation, we demonstrate that PuSHR dominates baselines ranging from purely decentralized to fully decentralized in terms of success rate and time efficiency across challenging tasks with up to 4 robots. Hardware experiments demonstrate the transfer of our system to the real world and highlight its robustness to model inaccuracies. Our code can be found at~\url{https://github.com/prl-mushr/pushr}, and videos from our experiments at~\url{https://youtu.be/DIWmZerF_O8}.
\end{abstract}

\section{Introduction}\label{sec:intro}

% We focus on the problem of nonprehensile rearrangement planning in tight spaces for a team of mobile robot pushers with non-holonomic constraints. In this problem, multiple robots must work together to modify a densely cluttered physical environment into a user-specified pattern via pushing, while avoiding collisions with each other. This is challenging because it requires parallel treatment of multirobot coordination, collision avoidance, and nonprehensile manipulation.

Multirobot systems have transformed sectors like fulfillment and warehouse automation. This was made possible in part thanks to efficient, scalable algorithms for multiagent pathfinding (MAPF)~\citep{sharon2015cbs,Kottinger2022cbs,chen2021pickupdelivery,hang2019lifelong} and task assignment (TA)~\citep{ecbsta,gerkey2004task-allocation,liu2019task,Qu2019MATA,korsah2013task-allocation}. Often, real-world robot deployment of such algorithms leverages extensive workspace and hardware engineering. Robots are typically holonomic, follow predefined paths (e.g., wire-guided), move in very constrained ways (e.g., rectilinearly like Amazon's Kiva robots~\citep{kiva}), and feature advanced gripping mechanisms. 

% introduction should be a brief summary of the results/discussion section.

% In this paper, we are driven by the challenge of tackling real-world rearrangement tasks with minimal hardware and workspace engineering, using non-specialized, inexpensive mobile robots. Leveraging pushing, mobile robots can be converted into effective mobile manipulators, capable of completing even challenging rearrangement tasks without requiring customized grippers. Using a team of nonholonomically constrained miniature robotic racecars~\citep{srinivasa2019mushr}, we focus on the task of rearranging a set of blocks into a specified planar pattern via pushing actions realized via their front bumper (see~\figref{fig:introfig}). The constraints of robots' kinematics, workspace boundary, collision avoidance, and pushing stability make this task challenging. 

\begin{figure}
    \centering
    \includegraphics[width=\linewidth]{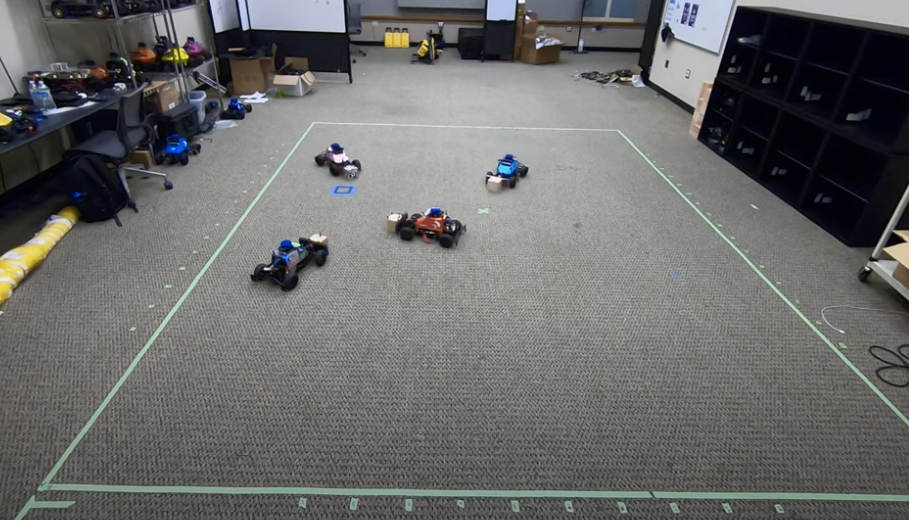}
    \caption{Using the PuSHR system, four MuSHR~\citep{srinivasa2019mushr} robots push four blocks toward their goals.}
    \label{fig:introfig}
\end{figure}

In this paper, we are driven by the challenge of tackling real-world rearrangement tasks on constrained hardware with minimal workspace engineering. Using a team of car-like robots~\citep{srinivasa2019mushr}, we consider the task of rearranging a set of cubic blocks into a desired planar pattern via pushing actions realized via their front bumper (see~\figref{fig:introfig}). Leveraging pushing, mobile robots can be converted into effective mobile manipulators, capable of completing even complex rearrangement tasks without requiring customized grippers. 

Simultaneously accounting for collision avoidance and push-stability in a tight space demands highly coordinated motion. Centralized MAPF algorithms can produce plans for complex rearrangement problems but accounting for model inaccuracies online requires frequent replanning that is impractical to execute on constrained hardware. On the other hand, decentralized controllers~\citep{ORCA,davis2019nh} provide scalability but lack an understanding of the global task structure which often gets them stuck on deadlocks.

% task structure into a global motion plan but directly tracking a MAPF plan on constrained hardware yields path deviations . This motivates additional considerations for the tracking controller such as collision avoidance and timing constraints. Balancing these considerations requires tedious tuning work and effectively increases task complexity. 

Our key insight is that optimally assigning pushing tasks to robots could simplify multi-robot coordination and enable robust performance via scalable decentralized control. Combined task and motion planning is intractable in our problem domain. Instead of simultaneously iterating over assignments and paths, we first plan an optimal task assignment considering a discretized workspace representation, and using this assignment, we plan collision-free trajectories for all robots in the continuous workspace. Each robot is then tracking its assigned trajectory via decentralized model predictive control accounting for collision avoidance and trajectory deviations. 

Through a simulation-based ablation study, we demonstrate that our system (PuSHR) is capable of handling a variety of complex rearrangement scenarios involving up to 4 robots, including asymmetric scenarios with more or fewer robots than objects. In particular, PuSHR is the only system that successfully completes all scenarios, often with the top time efficiency among a series of baselines ranging from baselines from fully decentralized to fully centralized control. Through hardware experiments on a team of MuSHR robots~\citep{srinivasa2019mushr}, we present statistical insights about the ability of PuSHR to handle model inaccuracies and produce robust performance across challenging scenarios with up to 4 robots.

\section{Related Work}

Our system design brings together multirobot system design, and both nonprehensile and collaborative manipulation. 

% \begin{figure*}
%   \dummyfig{Architecture} 
%   \caption{This should be some sort of a block diagram summarizing the whole system.}
%   \label{fig:architecture}
% \end{figure*}

\begin{figure*}
    \includegraphics[width=\linewidth]{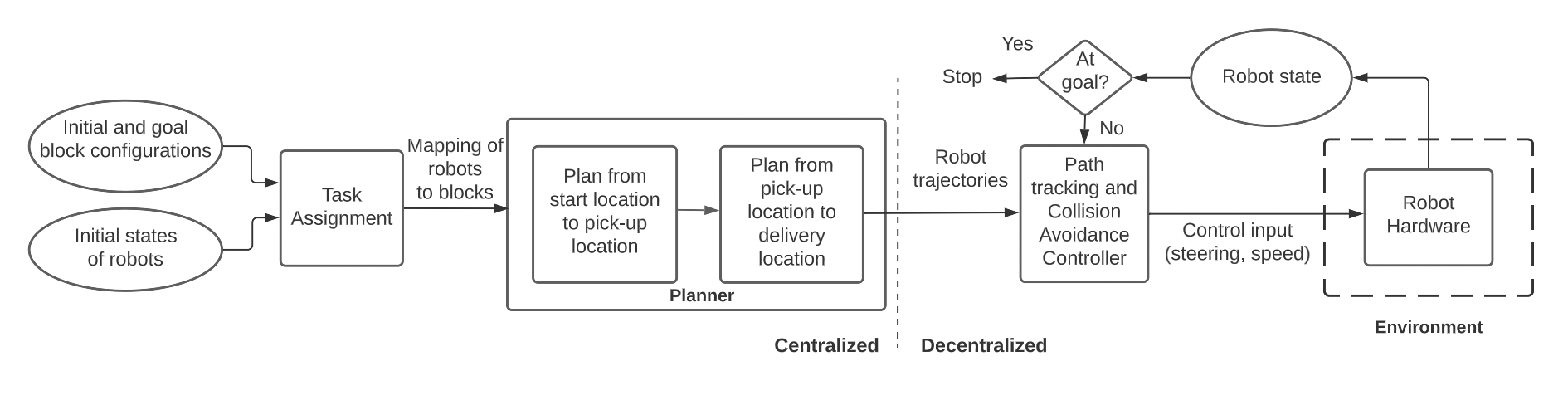}
    \caption{Overview of the PuSHR system architecture. Our system comprises task assignment, grasp planning, and local collision avoidance control. PuSHR guides a set of robots to push a set of blocks towards a set of goal configurations.} 
    \label{fig:architecture}
\end{figure*}

\subsection{Multirobot Systems}

Many important advances in MAPF came in recent years, driven by the scalability of multi-robot systems. Algorithms like $M^*$~\citep{wagner2011mstar} and Conflict Based Search (CBS)~\citep{sharon2015cbs} and their variants are behind many successful real-world deployments of multi-robot systems, ranging from fulfillment to warehouse and manufacturing. Such deployments have encouraged further investigation of MAPF variants like the multiagent pickup and delivery~\citep{liu2019task,hang2019lifelong,chen2021pickupdelivery} and the incorporation of kinodynamic constraints~\citep{clcbs,Kottinger2022cbs}. In parallel, TA has also found important applicability in physical domains involving teams of robots and/or humans~\citep{gerkey2004task-allocation,korsah2013task-allocation,Qu2019MATA}. In this paper, we integrate recent advances in kinodynamic MAPF~\citep{clcbs} and TA~\citep{ecbsta} into a real-world system.

Besides multi-robot planning, there is recent activity in control for dynamic multiagent domains. Graphics applications have motivated crowd simulators~\citep{ORCA}, whereas robot deployment in crowded and driving domains has led to proactive controllers for collision avoidance~\citep{mavrogiannis2018SM,roh2020corl}. The insights of such algorithms are often transferred in multi-robot domains where teams of robots navigate in close proximity~\citep{knepper2012multirobotca}. In this work, we leverage insights from multiagent collision avoidance~\citep{davis2019nh} into the design of a decentralized local controller that robustly adjusts for deviations from the multiagent plan during deployment.

% sartoretti2019RL

% multiagent pickup and delivery: The problem of multiple robots collaboratively taking "tasks" from an initial location to a final location is known as Multi-agent Pick-up and delivery. The usual way to do this is to use a heuristic based lower-bound cost estimate for each task assignment and then plan for the one with the lowest cost. In \citet{chen2021pickupdelivery}, \citet{ecbsta}, task assignment is informed by actual costs obtained from the planning methods for each task assignment instead of by heuristic based lower-bound estimates. This approach, while more accurate by virtue of using actual path costs, can be computationally expensive for agents with non-holonomic constraints. In our work, we propose using the latter approach to find just the task assignment. As the task assignment system uses a 2-D grid approximation of the workspace, the planning step is relatively quick and thus the optimal task assignment is found quickly. We then use a conflict based search (CBS) planner that considers the non-holonomic constraints of our agents to plan the actual paths given the task assignment. \citet{liu2019task} \citet{hang2019lifelong}

\subsection{Nonprehensile Manipulation}

Nonprehensile manipulation and especially push-based manipulation has been an active area of research~\citep{Stber2020LetsPT}. Early work studied mechanics and control for the quasistatic settings~\citep{Mason1986,goyalruina,howcutkosky,lynch1996}. Later work focused on planning using nonprehensile manipulation primitives for complex, cluttered environments~\citep{Dogar2012APF,king2013}. In ~\citep{multi-object-pushing}, they consider pushing multiple objects with a single pusher surface. Finally, some work has focused on handling parameter uncertainty by extracting data-driven force-motion models~\citep{Bauza2017,Zhou_IJRR18}, learning to adapt to different objects~\citep{Krivic2019}, or learning end-to-end pushing policies~\citep{yuan_ICRA18,pushnet}.
In this paper, we leverage insights on quasistatic pushing~\citep{howcutkosky,lynch1996,king2013} to inform the planning and low-level control components of our architecture.

\subsection{Collaborative Manipulation}

% Researchers have leveraged the scaleability of multirobot systems to distribute the execution of a manipulation task across teams of collaborating robots. 

An active line of work explores strategies of distributing manipulation tasks across a team of robots via implicit signaling encoded in robots' behaviors~\citep{mataric-multirobot-box-pushing,tsiamis2015,wang2016cooperative}. Some works developed multi-robot systems capable of assembling furniture~\citep{knepper2013ikeabot} or transporting deformable objects~\citep{mora2015collaborative} via prehensile manipulation. The assumption of rigidity at the contacts~\citep{tsiamis2015} often removes the need for force measurements~\citep{verginis2020}. This can be restrictive as contact parameters are challenging to estimate.
In some works robots act as sensors, collaboratively filtering important system parameters to guide the manipulation strategy~\citep{culbertson2021}.
Other works focus on alternative manipulation strategies such as multi-robot caging~\citep{fink2008multicaging} and pushing~\citep{mataric-multirobot-box-pushing}.

Unlike much of the prior work in collaborative manipulation which focuses on the transportation of a single object with holonomic pushers~\citep{mora2015collaborative,wang2016cooperative,culbertson2021, fish_box_push}, in this paper we consider the rearrangement of \emph{multiple} objects using nonholonomic pushers, with each object being controlled by only one pusher. 
%As our system only plans once
% We also employ a hybrid communication regime in which we perform much of the expensive computation (global planning, task assignment) in a centralized fashion but distribute trajectory tracking to individual robots.

\section{The Multiagent Nonprehensile Rearrangement Planning Problem}

% We focus on pushing of rectangular objects which from now on we will refer to as blocks. 

\begin{figure}
    \centering
    \includegraphics[width=\linewidth]{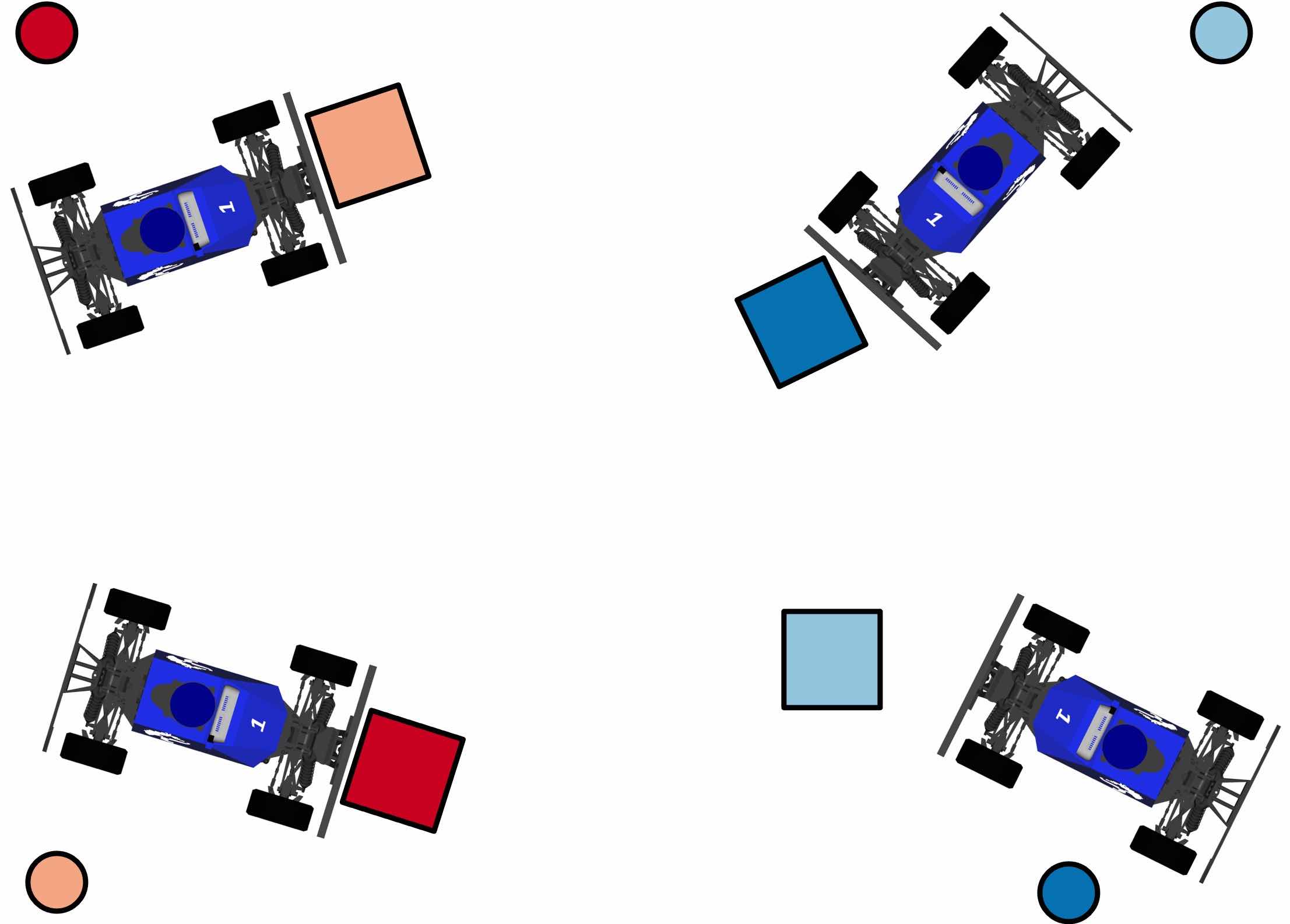}
    \caption{Problem setup. A team of mobile robots is pushing a set of blocks towards a set of goal locations, indicated as circles of same color. %We discretize the workspace to perform efficient assignment of blocks to robots.}
    \label{fig:setup}}
\end{figure}

We consider a team of $n$ mobile robots and a set of $m$ rectangular blocks, embedded in a workspace $\mathcal{W}\subseteq\mathbb{R}^2$. We denote the states of the robots as $p^i\in SE(2)$, $i\in\mathcal{N} = \{1,\dots,n\}$ and the states of the objects as $o^j\in \mathbb{R}^2$, $j\in\mathcal{M} = \{1,\dots, m\}$ (we ignore objects' orientation). Each robot $i$ follows rear-axle simple-car kinematics $p^i_{k+1} = f(p^i_{k}, u^i_{k})$, where $u^i_{k}$ represents a control action, drawn from a space of velocities and steering angles $\mathcal{U} = [-v_{max}, v_{max}] \times [-\phi_{max}, \phi_{max}]$ at timestep $k$. Robots may manipulate objects via pushing realized via flat-surface bumpers attached at their front (see~\figref{fig:setup}). The goal of the robots is to rearrange the objects from an initial configuration $S = (s^1,\dots,s^m)$ to a goal configuration $G = (g^1,\dots,g^m)$ via sequences of push-based manipulation actions. Our goal is to design a system to enable the robots to successfully complete the rearrangement from $S$ to $G$ in a time-efficient manner that scales robustly with the number of robots. %\msnote{not sure if it is necessary but time-efficient could be converted to math. E.g. sum of distances all the robots travel.}

\section{PuSHR: A Multirobot System for Nonprehensile Rearrangement}\label{sec:pushr}

We describe a planning architecture for multi-robot, multi-object nonprehensile rearrangement problems. 

\subsection{Architecture Overview}\label{sec:architecture}

As shown in~\figref{fig:architecture}, our architecture consists of three main layers: a) a centralized task-assignment module that assigns objects to robots; b) a centralized global planning module that assigns paths to robots; c) a decentralized control module that enables robots to track their paths while maintaining contact with their assigned objects and avoiding collisions. The process begins with the task assignment system receiving the initial location of the robots and the pickup and delivery location of the tasks. The task assignment system then provides the global planner with the optimized robot-task pairs. The global planner produces a space-time trajectory for each robot. The onboard local planner then produces the controls to follow said trajectory. %\cmnote{here, we just need to briefly describe the interfaces between the components of the architecture.}

\subsection{Task Assignment}

Given robots' initial states $P_0=(p^1_0,\dots,p^n_0)$ and blocks' initial and final configurations, $S$ and $G$ respectively, the objective of the task assignment module is to generate an efficient assignment of robots to blocks, i.e., $\mathcal{N}\to\mathcal{M}$. Efficiency refers to the total distance required for all robots to complete their rearrangement tasks. We cast this problem as an instance of multiagent pathfinding (MAPF) on a discrete graph and solve it using the Enhanced Conflict Based Search with Task Assignment (ECBS-TA) by~\citet{ecbsta} which we adapt to our problem domain below.

For computation considerations, we use a discretized representation of the workspace for task assignment. We partition the workspace $\mathcal{W}$ into a set of discrete regions (see~\figref{fig:setup}) whose connectivity we describe as a graph $\mathcal{G} = (V,E)$, where vertices $v\in V$ represent regions, and edges $e\in E$ are adjacency relationships between them. Based on this partition, we map the state of a robot/block to the workspace region that contains the majority of its volume. We assume that at each timestep, a robot can only move to an adjacent vertex or wait at its current vertex. %We denote by $v_t^i$ the vertex of agent $i$ at time $t$.
%\ssnote{I don't get it. A vertex is in R2 whereas a robot is in SE(2)}

% \cmnote{do we handle cases where $m>n$ or $m<n$? If so, we need to say how we do that.}

% We invoke ECBS-TA to jointly search for short, collision-free paths for all robots. 

ECBS-TA searches for a collision-free path set $\mathcal{P} = (\mathcal{P}^1,\dots, \mathcal{P}^n)$ such that path $i$ connects the initial region of robot $i\in\mathcal{N}$ with the initial and final region of block $j\in\mathcal{M}$. The search sequentially refines path assignments by resolving emerging robot collisions, driven by a Euclidean-distance-based heuristic until a collision-free path set $\mathcal{P}$ is found. A byproduct of this process is the assignment of block $j\in\mathcal{M}$ to robot $i\in\mathcal{N}$ $\forall i\in\mathcal{N}$. We modify ECBS-TA to handle problems with $n\neq m$ as follows: when $n>m$, the $m$ best robots complete the rearrangement, and when $n<m$, the robots rearrange the $n$ most convenient blocks first, then task assignment is invoked again to assign the remaining $m-n$ blocks to robots until all blocks are assigned.

\subsection{Centralized Global Planning}\label{sec:globalplanning}

The paths produced by the task assignment module do not account for robot kinematics. Thus, we invoke a global planner to generate smooth, collision-free robot trajectories that respect the task assignment captured in $\mathcal{P}$. For this, we use CL-CBS~\citep{clcbs}, an algorithm that searches for a collision-free trajectory set $\mathcal{R} = (\mathcal{R}^1,\dots,\mathcal{R}^n)$, where $\mathcal{R}^i = (r^i_0,\dots,r^i_K)$ is the trajectory planned for robot $i\in\mathcal{N}$. The search process assembles individual trajectories from a set of kinematically feasible motion primitives and iterates until conflicts are resolved using a conflict-based search (CBS) methodology~\citep{sharon2015cbs}. These primitives are: stopping in place; moving forward/backward by 1 unit (equal to speed of the robot multiplied by the timestep); moving forward left/right on a unit-length circular arc; moving backward left/right on a unit-length circular arc.

We invoke CL-CBS twice: first to plan a set of paths that take robots from $P_0$ to $S$, and then to take the robots from $S$ to $G$. The output of this process is a time-indexed trajectory set $\mathcal{R} = (\mathcal{R}^1,\dots,\mathcal{R}^n)$ where $\mathcal{R}^i = (r^i_0,\dots,r^i_K)$ is the trajectory planned for robot $i\in\mathcal{N}$.

\begin{figure*}
    \centering
    \includegraphics[width=\linewidth]{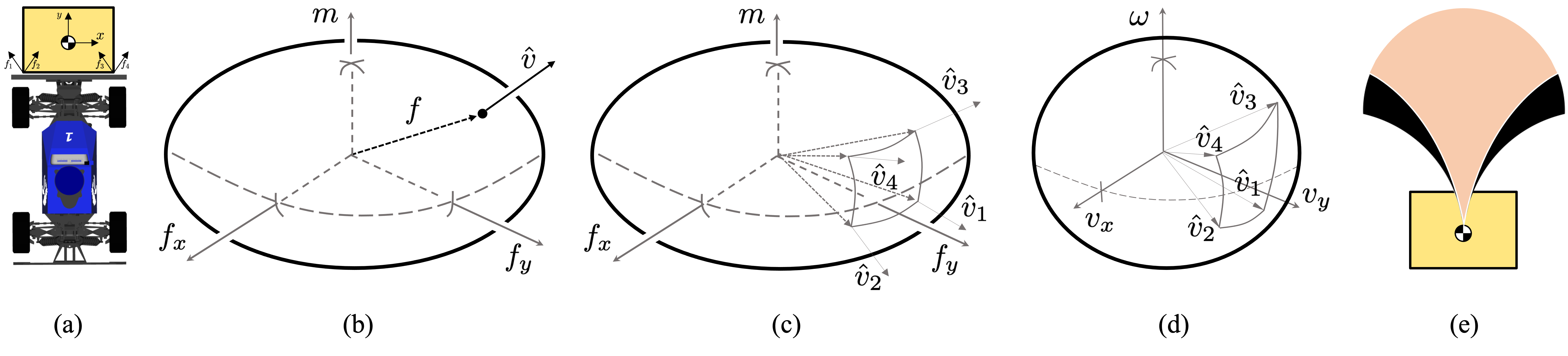}
    \caption{Deriving the stable set of controls for quasistatic pushing following prior work of~\citep{lynch1996,goyalruina,king2013}. (a) Robot-block pushing system, and static friction boundary forces. (b) Limit surface, mapping force on the object to resulting unit velocity. (c) Boundary friction forces are mapped to boundary unit velocities through the limit surface. (d) The stable set of unit velocities represented on the unit control sphere. (e) Full (black color) and stable (beige) set of controls mapped to block velocities.}
    \label{fig:pushing}
\end{figure*}

\subsection{MPC for Multiagent Collision Avoidance}\label{sec:mpc}

Even small inconsistencies in following the timing of the planned trajectories may accumulate and result in collisions when directly following $\mathcal{R}$. For this reason, we deploy a decentralized model predictive controller (MPC) on each robot that besides path tracking, accounts for multi-agent collision avoidance. We formulate this problem for robot $i$ as a discrete optimization over a set of control trajectories $\bm{\mathcal{U}}$. For robot $i$, at every loop, the MPC outputs a trajectory $\boldsymbol{u}^{i*}$ determined as follows:

\begin{equation}
\begin{split}
\boldsymbol{u}^{i*} = \arg  \min_{\boldsymbol{u}^i\in\boldsymbol{\mathcal{U}}} \sum_{k=0}^{N-1} &\Big(a_{cte}\mathcal{J}_{cte}(p^i_k,r^i_k) + a_{col}\mathcal{J}_{col}(P_k)\\ &+ a_{time}\mathcal{J}_{time}(p^i_k,r^i_k)\Big)\\
    s.t.\: & p^i_{k+1} = f(p^i_k, u^i_k)
   \label{eq:mpc}
\end{split}\mbox{,}
\end{equation}
where $\mathcal{J}_{cte}$ is a cross-track error cost forcing the robot to stay close to its reference path, $\mathcal{J}_{col}$ is a collision avoidance cost penalizing proximity between robots, $\mathcal{J}_{time}$ is a timing tracking cost, penalizing deviations from the timing defined by the global plan, and $a_{cte}$, $a_{col}$,$a_{time}$ are weights. The timing cost is defined as follows:
\begin{equation}
    % \begin{split}
        % \mathcal{J}_{cte}(p_k^i,r_k^i) = & W_{cte} * {cte}^i_k\\
        \mathcal{J}_{time}(p_k^i,r_k^i) =  \widehat{ate}^i(t) - {ate}^i_k
    % \end{split}
\end{equation}
where $\widehat{ate}^i$ is the reference along-track error from a waypoint as a function of time resulting from the robot's planned path, and ${ate}^i_k$ is the along-track error for a given time-step. The collision cost is defined as:

\begin{equation}
    % \begin{split}
        \mathcal{J}_{col}(p_k) =  \sum_{j\in\mathcal{N}\backslash i}\max({d}_{thr} - ||p^k_i - p^k_j||,0)
\label{eq:collisioncost}
\end{equation}
where $d_{thr}$ is a threshold beyond which the cost is 0. 

%$W_{cte}, W_{time}, W_{col}$ refer to the cross track error weight, the timing error weight and the collision weight respectively.

% \begin{figure}
%     \centering
%     \includegraphics[width = \linewidth]{figures/pushing.png}
% etup    \caption{REPLACE WITH MULTIAGENT MULTIOBJECT PUSHING s}
%     \label{fig:pushing}
% \end{figure}

\subsection{Stable Pushing}\label{sec:stablepushing}
% \msnote{you need to cite where you are getting this process from}

% Our system relies on pushing to drive objects to their goal locations (see~\figref{fig:combined-fc-stable-set}).
% To push an object towards a desired location, the robot needs to maintain controllability of the object throughout its path, which is challenging due to the uncertainty over contact parameters~\citep{lynch1996}. 

 %To do that, we follow the methodology found in literature on pushing models and systems~\citep{Mason1986,goyalruina,lynch1996,howcutkosky,king2013,Zhou_IJRR18}.

% We only model the contact between a robot's flat bumper and an object (see~\figref{fig:pushing}), ignoring possible object-object and robot-robot contacts. 

Following~\citet{lynch1996}, assuming quasistatic pushing at the block/bumper contact, we derive the stable set of unit velocities $\mathcal{\hat{U}}_{stable}$ that guarantee that the pushed object remains fixed on the bumper. We model the contact between the robot's bumper and a block as a line defined by two point contacts at the edges of the block (\figref{fig:pushing}a). Given the block/bumper friction coefficient $\mu$, we calculate the boundary forces $f_1, f_2, f_3, f_4$ and use them to construct a composite friction cone. Through the limit surface~\citep{goyalruina,king2013} ((see~\figref{fig:pushing}b), we map these forces to limit unit velocities $\hat{v}_1, \hat{v}_2, \hat{v}_3, \hat{v}_4$, representing the boundary of the stable set $\mathcal{\hat{U}}_{stable}$ (\figref{fig:pushing}c), beyond which the block would start sliding. Finally, we map these velocities to limits of robot controls using the robot kinematics (\figref{fig:pushing}d) and extract a limiting turning radius that the robots must respect to maintain control of the blocks while pushing (\figref{fig:pushing}e). Assuming a coefficient of friction $\mu \approx 0.6$ (which we experimentally measured upon attaching sandpaper on the robots' bumpers), we found this radius to be $1.6 m$, corresponding to a steering-angle limit of $\mathcal{\phi}_{max} = 0.17 rad$ for a MuSHR robot~\citep{srinivasa2019mushr}. This value was also confirmed through experiments for speeds ranging from $0.3-0.8 m/s$.

\subsection{Implementation}\label{sec:implementation}

\begin{figure*}
    \centering
    \begin{subfigure}{.115\linewidth}
        \centering
        \includegraphics[width = \linewidth]{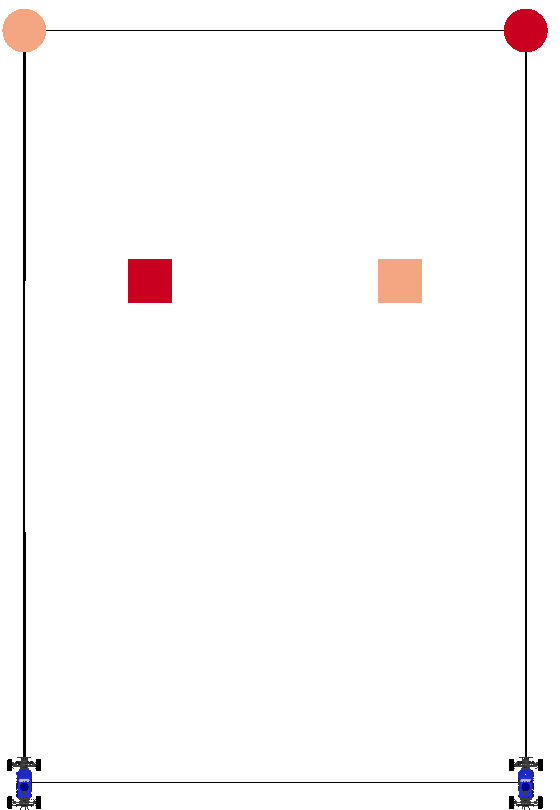}
        \caption{ S2.a.\label{fig:scenario2a}}
    \end{subfigure}
    \begin{subfigure}{.115\linewidth}
        \centering
        \includegraphics[width = \linewidth]{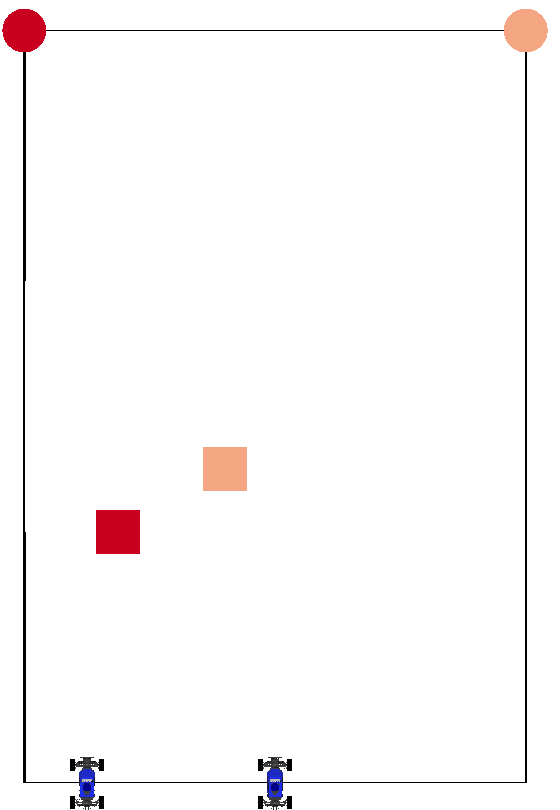}
        \caption{ S2.b.\label{fig:scenario2b}}
    \end{subfigure}
    \begin{subfigure}{.115\linewidth}
        \centering
        \includegraphics[width = \linewidth]{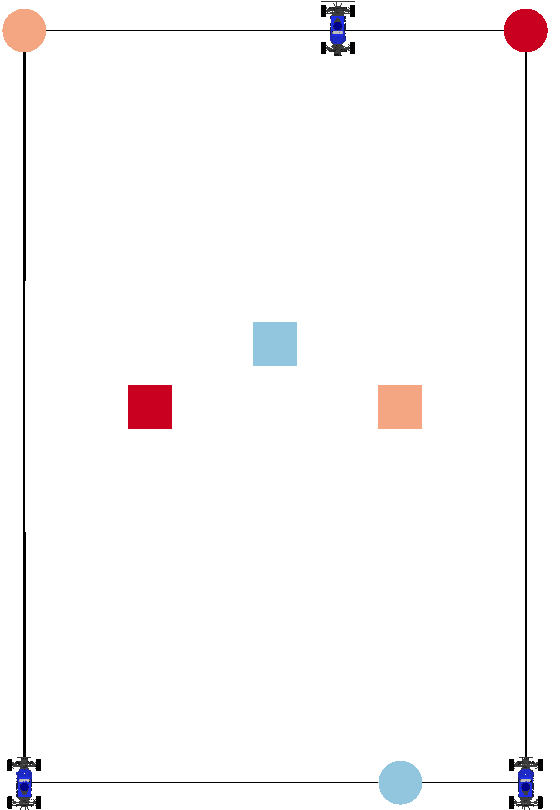}
        \caption{ S3.a.\label{fig:scenario3a}}
    \end{subfigure}    
    \begin{subfigure}{.115\linewidth}
        \centering
        \includegraphics[width = \linewidth]{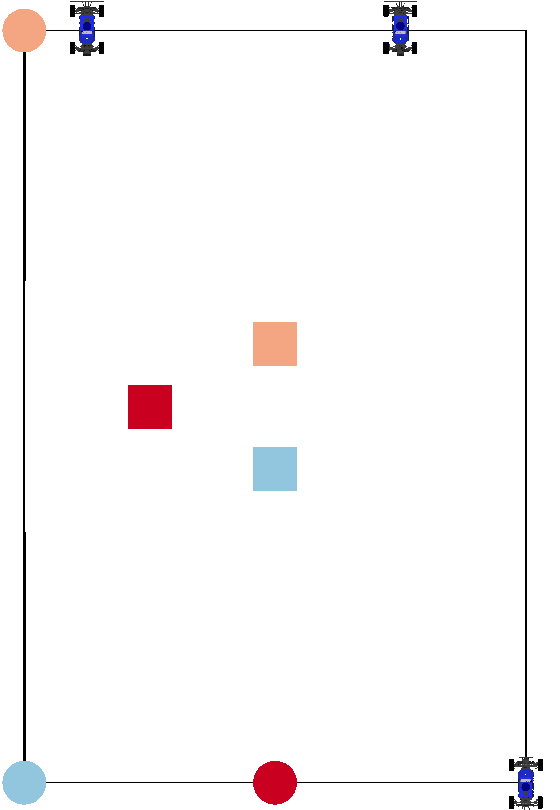}
        \caption{ S3.b.\label{fig:scenario3b}}
    \end{subfigure}  
    \begin{subfigure}{.115\linewidth}
        \centering
        \includegraphics[width = \linewidth]{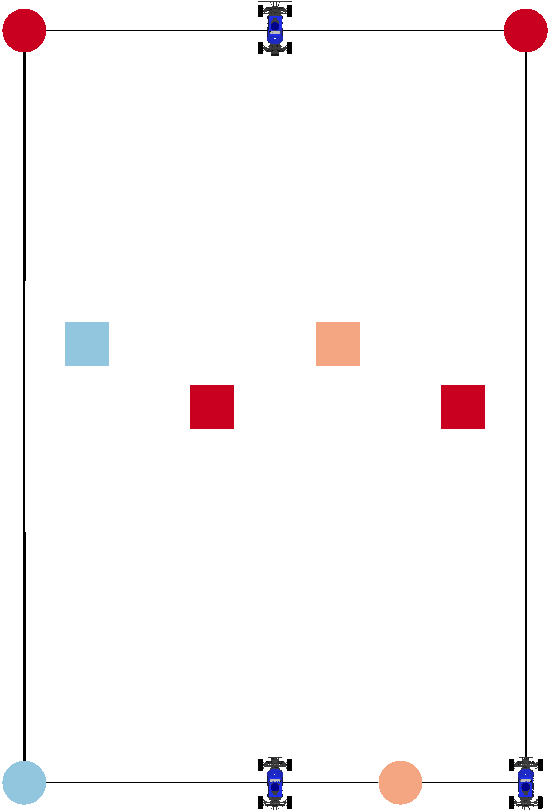}
        \caption{ S3.c.\label{fig:scenario3c}}
    \end{subfigure}    
    \begin{subfigure}{.115\linewidth}
        \centering
        \includegraphics[width = \linewidth]{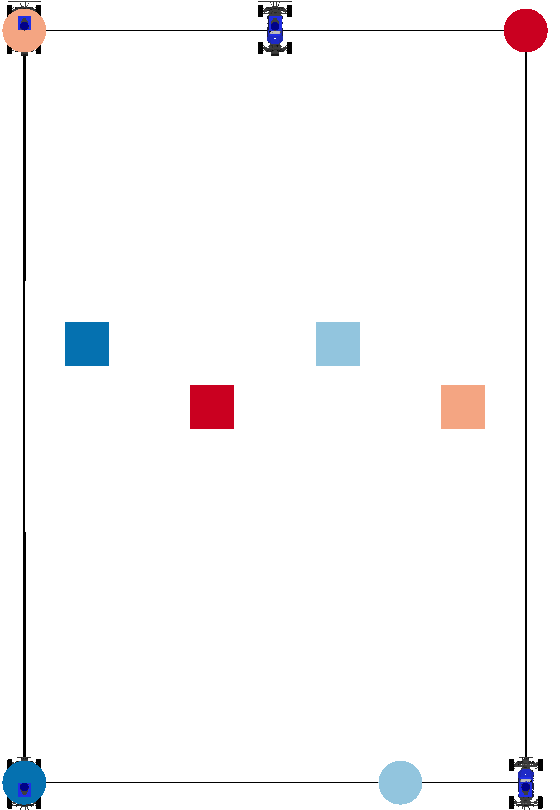}
        \caption{ S4.a.\label{fig:scenario4a}}
    \end{subfigure}  
    \begin{subfigure}{.115\linewidth}
        \centering
        \includegraphics[width = \linewidth]{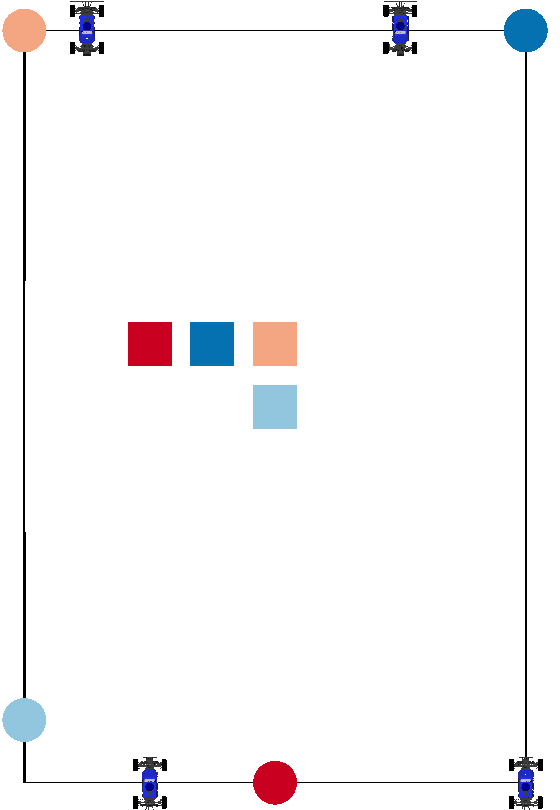}
        \caption{ S4.b.\label{fig:scenario4b}}
    \end{subfigure}    
    \begin{subfigure}{.115\linewidth}
        \centering
        \includegraphics[width = \linewidth]{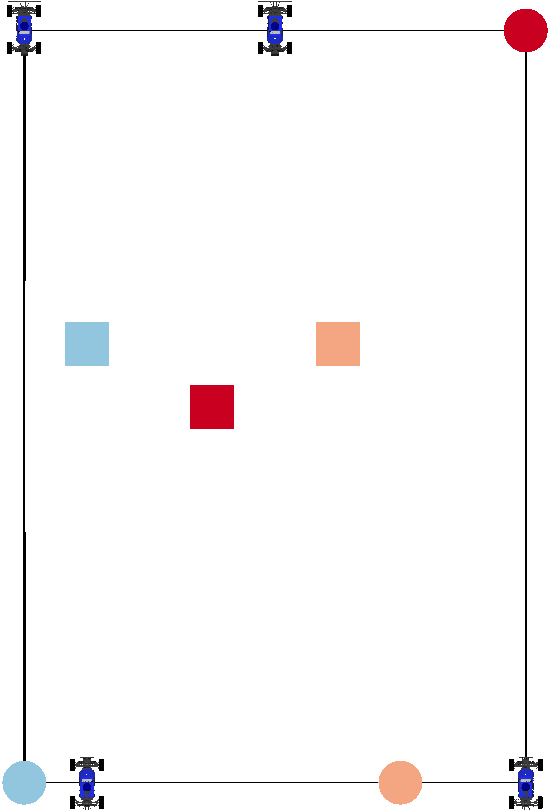}
        \caption{ S4.c.\label{fig:scenario4c}}
    \end{subfigure}  
    \caption{Scenarios used in our evaluation. The cars are shown in their initial configurations. Initial block configurations are shown as squares whereas their goal positions are shown as circles (we ignore orientations). \label{fig:scenarios}}
\end{figure*}

In our ECBS-TA implementation, we set the suboptimality factor $w = 1.3$, which gave high-quality assignments within acceptable times. We modified CL-CBS~\citep{clcbs} to account for pushing stability: when planning a pushing trajectory phase, we restrict the maximum steering angle to be less than $\mathcal{\phi}_{max}$. Our MPC used $d_{thr} = 0.6 m$, $a_{cte} = 200$, $a_{time} = 20$, $a_{col} = 15$, which we obtained via parameter sweeps over success rate and minimum distance for scenarios similar to S3.a,b, S4.a,b (see~\figref{fig:scenarios}). To practically remain within the quasistatic regime, we set $v_{max} = 0.4 m/s$, $\phi_{max} = 0.314\ rad$ during non-pushing phases, and $\phi_{max} = 0.17\ rad$ during pushing phases.

\begin{table*}
\centering
% \small
\caption{Success rate over 100 simulated trials for each scenario and algorithm.
\label{tab:success}
}
\resizebox{\linewidth}{!}{%
\addvbuffer[0pt 0pt]{
\begin{tabular}{l|cccccccc}
\toprule
\multicolumn{1}{l}{\textbf{Configuration}} &
  \multicolumn{2}{c}{2 robots - 2 objects} &
  \multicolumn{2}{c}{3 robots - 3 objects} &
  \multicolumn{1}{c}{3 robots - 4 objects} &
  \multicolumn{2}{c}{4 robots - 4 objects} &
  \multicolumn{1}{c}{4 robots - 3 objects} \\ \midrule
\multicolumn{1}{l}{\textbf{Scenario}} &
  S2.a &
  \multicolumn{1}{c|}{S2.b} &
  S3.a &
  \multicolumn{1}{c|}{S3.b} &
  \multicolumn{1}{c|}{S3.c} &
  S4.a &
  \multicolumn{1}{c|}{S4.b} &
  S4.c \\ \midrule 
\multicolumn{1}{l|}{LC} &
  \textcolor{black}{0.99} &
  \multicolumn{1}{c|}{{0.00}} &
  \textcolor{black}{0.00} &
  \multicolumn{1}{c|}{\textcolor{black}{0.00}} &
  \multicolumn{1}{c|}{\textcolor{black}{0.00}} &
  \textcolor{black}{0.00} &
  \multicolumn{1}{c|}{\textcolor{black}{0.00}} &
  \textcolor{black}{0.00} \\
\multicolumn{1}{l|}{TA-LC} &
  \textcolor{black}{0.99} &
  \multicolumn{1}{c|}{{0.99}} &
  \textcolor{black}{0.89} &
  \multicolumn{1}{c|}{\textcolor{black}{0.99}} &
  \multicolumn{1}{c|}{\textcolor{black}{0.00}} &
  \textcolor{black}{0.98} &
  \multicolumn{1}{c|}{\textcolor{black}{0.00}} &
  \textcolor{black}{0.73} \\
\multicolumn{1}{l|}{GP} &
  \textcolor{black}{0.99} &
  \multicolumn{1}{c|}{\textcolor{black}{0.98}} &
  \textcolor{black}{0.88} &
  \multicolumn{1}{c|}{\textcolor{black}{1.00}} &
  \multicolumn{1}{c|}{\textcolor{black}{0.00}} &
  \textcolor{black}{0.99} &
  \multicolumn{1}{c|}{\textcolor{black}{0.00}} &
  \textcolor{black}{0.14} \\
\multicolumn{1}{l|}{GP-CA} &
  \textcolor{black}{1.00} &
  \multicolumn{1}{c|}{\textcolor{black}{0.98}} &
  \textcolor{black}{0.82} &
  \multicolumn{1}{c|}{\textcolor{black}{0.99}} &
  \multicolumn{1}{c|}{\textcolor{black}{0.00}} &
  \textcolor{black}{0.96} &
  \multicolumn{1}{c|}{\textcolor{black}{0.20}} &
  \textcolor{black}{0.92} \\
\multicolumn{1}{l|}{PuSHR} &
  \textcolor{black}{\textbf{1.00}} &
  \multicolumn{1}{c|}{\textcolor{black}{\textbf{0.99}}} &
  \textbf{0.99} &
  \multicolumn{1}{c|}{\textcolor{black}{\textbf{1.00}}} &
  \multicolumn{1}{c|}{\textbf{0.96}} &
  \textcolor{black}{\textbf{0.99}} &
  \multicolumn{1}{c|}{\textbf{{1.00}}} &
  \textbf{0.98} \\
\bottomrule
\end{tabular}
}
}
\end{table*}

\section{Evaluation}

We conduct simulated and real-world experiments using MuSHR~\citep{srinivasa2019mushr} robots deployed in a workspace of size $4\times 6 m^2$.

%to evaluate our system and characterize its scaleability and robustness. For our experiments, we use

\subsection{Experiment Design}\label{sec:experiment}

\textbf{Scenarios}: We investigate the performance of our system across 8 scenarios of varying difficulty involving two, three, and four robots (see~\figref{fig:scenarios}). The scenarios involve navigating from a set of initial robot configurations towards a set of initial block configurations, making contact with the blocks, and pushing them toward the blocks' goal configurations. They were designed to give rise to challenging encounters among robots.

%The scenarios are designed as follows: We use the two robot experiments as sanity checks. All systems should be able to complete these test cases. In these cases, the delivery point is on one side of the pick up point and the robot on the other, meaning that the robot does not need to execute complex maneuvers like u-turns before or after picking the block to reach the end destination.

\textbf{Algorithms}: To demonstrate the value of PuSHR, we perform an ablation study, comparing its performance against a set of variants. Note that in the following descriptions, \emph{manual TA} refers to TA that ranks around the median of all possible assignments ordered with respect to the time cost used by ECBS-TA~\citep{ecbsta}. See~\figref{fig:taPathFeatureComparison} for an example of how different assignments in scenario S4b rank and map to path planning Makespan.

\textit{LC}. We implemented a Local Control (LC) baseline using the NonHolonomic Time To Collision (NHTTC)~\citep{davis2019nh}, an optimization-based decentralized controller that accounts for robot kinematics and multi-robot collision avoidance. NHTTC evaluates a set of constant-radius trajectories with respect to distance to goal and time to collision. During pushing, we constrain these trajectories to be in the stable set (see~\figref{fig:pushing}e). This baseline uses \emph{manual} TA.
% LC uses manual task assignment based on the following rule: $n-1$ robots are assigned tasks such that the block's goal and the robot's start are on the same side of the block's start location. 
%This is done to emphasize the difficulty of navigating in tight-spaces. 
%For manual assignment of tasks, in scenarios with 2 robots, the robots and the delivery locations would be on opposite sides of the pick-up location, and the assignment would force the robots to cross paths. 

\textit{TA-LC}. This baseline is also LC, executed considering \emph{optimal TA} provided by ECBS-TA~\citep{ecbsta}.

\textit{GP}. Using \emph{manual} TA, the Global Planning (GP) baseline invokes CL-CBS~\citep{clcbs} to plan trajectories for all robots, and a decentralized MPC to track the trajectories. This MPC is equivalent to the description of eq.~(\ref{eq:mpc}) with $a_{col}=0$.

% 3. \textbf{GP}. Our Global Planning (GP) baseline uses manual block assignment (same as LC); then invokes CL-CBS~\citep{clcbs} to plan global trajectories for all robots, tracked using a local controller implemented as a version of the MPC from eq.~\ref{eq:mpc} but \emph{without the collision-avoidance cost}.

% The CLCBS planner produces a space time trajectory for each robot which is then tracked by a pure-pursuit controller.
\textit{GP-CA}. This baseline is identical to GP but uses $a_{col}\neq 0$ in eq.~(\ref{eq:mpc}) to account for collision avoidance.

% 4. \textbf{GP-CA}. The Global Planning with Collision Avoidance (GP-CA) baseline involves manual block assignment, and a set of global plans provided by CL-CBS. These plans are then tracked using the full version of the MPC described in Sec.~\ref{sec:mpc} which \emph{accounts for local collision avoidance}.

\textit{PuSHR}. PuSHR is identical to GP-CA, but leverages \emph{optimal TA} through ECBS-TA~\citep{ecbsta} as described in Sec.~\ref{sec:pushr}.

% 5. \textbf{TA-GP-CA}: Our complete system involves \emph{Optimal Task Assignment with Global Planning and Collision Avoidance} (TA-GP-CA), as discussed in~\ref{sec:architecture}. The task assignment module provides an optimal assignment of blocks to robots; then CL-CBS is invoked to plan trajectories to the blocks, and from the blocks to the block goals. These trajectories are tracked independently by each robot's MPC.

\textbf{Metrics}:
We evaluate performance with respect to:

\emph{Success rate}. We consider a trial to be successful if all robots are able to move their assigned blocks within a threshold distance of $0.1 m$ from their goals. Missing the block is also considered a failure even if the robots make it to their end location within the specified tolerance. %The success rate for an algorithm is the ratio of successful trials over the total number of trials.

%robots reach their final destination without colliding with each other and without missing the block at the pick-up location as the car must be within some tolerance of the block's pick up location to actually pick the block.

\emph{Makespan}. The time taken by the last robot to reach the goal of its assigned block in a successful trial. The clock starts when the plans are published by the global planner.%, and so the Makespan does not include the time taken by the task assignment and the global planner.

\emph{Minimum distance}. The minimum Euclidean distance between any two robots during a successful trial. A large minimum distance is advantageous as it provides an additional buffer to correct plan deviations and avoid collisions. 

\textbf{Hypotheses}: We generally expect that the spatial structure representation introduced through TA and GP will improve the system's time efficiency. We also expect that the collision avoidance (CA) module will enable robots to keep a larger buffer between each other. We distill our expectations about system performance into the following hypotheses:

\textit{H1}. GP achieves higher success rate than LC.

\textit{H2}. GP-CA achieves higher minimum distance than GP.

% \textbf{H3}. TA-LC and TA-GP-CA achieve higher success rates and lower Makespans than their non-TA versions, i.e., LC and GP-CA respectively.

\textit{H3}. TA improves the success rate for LC and GP-CA.

\textit{H4}. PuSHR is the most successful across all scenarios.

\textit{H5}. PuSHR maintains block control in the real world.

\textbf{Experimental Process}. To extract statistical insights about the \emph{planning} and \emph{tracking} performance of our system, we ran a series of scenarios (see~\figref{fig:scenarios}) in simulation. For each scenario, we generated 100 different trials, each involving a random spatial perturbation (radius $0.05 m$) of all robots' initial configurations. These perturbations allow us to understand the robustness of our system while ensuring the consistency of the scenarios being tested. We executed all trials for each scenario with each of the algorithms considered. 

%Note that while our simulations did not include physics modeling, our simulated trials are designed to be a indicative of our system's ability to securely push objects: having small cross-track error at the pickup location guarantees contact with the block. Second, given that our controls upon contact are bounded within the stable pushing set, the block is guaranteed to stay in contact throughout the pushed portion of the trajectory. 

%For the pushing model of Sec.~\ref{sec:stablepushing}, we assumed a block-bumper friction coefficient $\mu=0.6$, which was the value we measured in lab experiments upon attaching a piece of sandpaper on the bumper.

% \atnote{Due to the constrained nature of robot kinematics during the stable pushing phase and the relatively small size of our workspace, testcases with randomly sampled robot initial states and block initial and final configurations would have a high probability of being kinematically infeasible}.

% \atnote{May want to explain that this is because of tight workspace, if we e.g. randomly generated testcases many would not be kinematically feasible}

To characterize the robustness of PuSHR to model inaccuracies, we executed some of the more challenging scenarios in an identical lab setup. As blocks, we used a set of wooden cubes with a side of $0.1m$. Our workspace was fully covered by a motion capture system of 12 overhead cameras, providing high-accuracy localization of robots ($\sim 1mm$). Note that we only recorded the initial block configurations, letting our system push the blocks to their goals by leveraging the stable pushing derivation, without closing the loop on block position.

% \atnote{Give some more context to why this is ok -- maybe we don't need to model pushing here because being having sufficiently small cross track error at pickup will guarantee good contact and block retention, as demonstrated in lab experiments}

% \cmnote{VERIFY THESE TRENDS WHEN WE GET THE FINAL DATA}

\begin{table*}
\centering
\caption{Makespan ($s$) over 100 simulated trials for each scenario and algorithm.
\label{tab:makespan}
}
\resizebox{\linewidth}{!}{%
\addvbuffer[0pt 0pt]{
\begin{tabular}{l|cccccccc}
\toprule
\multicolumn{1}{l}{\textbf{Configuration}} &
  \multicolumn{2}{c}{2 robots - 2 objects} &
  \multicolumn{2}{c}{3 robots - 3 objects} &
  \multicolumn{1}{c}{3 robots - 4 objects} &
  \multicolumn{2}{c}{4 robots - 4 objects} &
  \multicolumn{1}{c}{4 robots - 3 objects} \\ \midrule
\multicolumn{1}{l}{\textbf{Scenario}} &
  S2.a &
  \multicolumn{1}{c|}{S2.b} &
  S3.a &
  \multicolumn{1}{c|}{S3.b} &
  \multicolumn{1}{c|}{S3.c} &
  S4.a &
  \multicolumn{1}{c|}{S4.b} &
  S4.c \\ \midrule 
\multicolumn{1}{l|}{LC} &
  \textcolor{black}{29.25 $\pm$ 2.83} &
  \multicolumn{1}{c|}{{N/A}} &
  \textcolor{black}{N/A} &
  \multicolumn{1}{c|}{\textcolor{black}{N/A}} &
  \multicolumn{1}{c|}{\textcolor{black}{N/A}} &
  \textcolor{black}{N/A} &
  \multicolumn{1}{c|}{\textcolor{black}{N/A}} &
  \textcolor{black}{N/A} \\
\multicolumn{1}{l|}{TA-LC} &
  \textcolor{black}{30.03 $\pm$ 2.97} &
  \multicolumn{1}{c|}{{25.03 $\pm$ 0.00}} &
  \textbf{17.02 $\pm$ 0.00} &
  \multicolumn{1}{c|}{\textbf{17.91 $\pm$ 0.31}} &
  \multicolumn{1}{c|}{\textcolor{black}{N/A}} &
  \textbf{17.60 $\pm$ 0.49} &
  \multicolumn{1}{c|}{\textcolor{black}{N/A}} &
  \textcolor{black}{\textbf{14.17 $\pm$ 0.36}} \\
\multicolumn{1}{l|}{GP} &
  \textcolor{black}{20.02 $\pm$ 0.10} &
  \multicolumn{1}{c|}{\textcolor{black}{32.75 $\pm$ 0.52}} &
  \textcolor{black}{33.84 $\pm$ 0.44} &
  \multicolumn{1}{c|}{\textcolor{black}{29.06 $\pm$ 0.01}} &
  \multicolumn{1}{c|}{\textcolor{black}{N/A}} &
  \textcolor{black}{34.08 $\pm$ 0.02} &
  \multicolumn{1}{c|}{\textcolor{black}{N/A}} &
  \textcolor{black}{\textcolor{black}{32.49 $\pm$ 0.50}} \\
\multicolumn{1}{l|}{GP-CA} &
  \textcolor{black}{20.03 $\pm$ 0.00} &
  \multicolumn{1}{c|}{\textcolor{black}{32.72 $\pm$ 0.49}} &
  \textcolor{black}{33.98 $\pm$ 0.50} &
  \multicolumn{1}{c|}{\textcolor{black}{30.12 $\pm$ 0.83}} &
  \multicolumn{1}{c|}{\textcolor{black}{N/A}} &
  \textcolor{black}{34.34 $\pm$ 0.44} &
  \multicolumn{1}{c|}{\textcolor{black}{31.20 $\pm$ 0.35}} &
  \textcolor{black}{\textcolor{black}{32.76 $\pm$ 0.46}} \\
\multicolumn{1}{l|}{PuSHR} &
  \textbf{20.02 $\pm$ 0.00} &
  \multicolumn{1}{c|}{\textcolor{black}{\textbf{13.02 $\pm$ 0.00}}} &
  \textcolor{black}{22.01 $\pm$ 0.17} &
  \multicolumn{1}{c|}{\textcolor{black}{20.61 $\pm$ 0.50}} &
  \multicolumn{1}{c|}{\textbf{51.55 $\pm$ 0.85}} &
  \textcolor{black}{23.36 $\pm$ 0.47} &
  \multicolumn{1}{c|}{\textbf{{21.05 $\pm$ 0.01}}} &
  \textcolor{black}{\textcolor{black}{18.03 $\pm$ 0.01}} \\
\bottomrule
\end{tabular}
}
}
\end{table*}

\begin{table*}
\centering
\caption{Minimum distance ($m$) over 100 simulated trials for each scenario and algorithm.
\label{tab:minDist}
}
\resizebox{\linewidth}{!}{%
\addvbuffer[0pt 0pt]{
\begin{tabular}{l|cccccccc}
\toprule
\multicolumn{1}{l}{\textbf{Configuration}} &
  \multicolumn{2}{c}{2 robots - 2 objects} &
  \multicolumn{2}{c}{3 robots - 3 objects} &
  \multicolumn{1}{c}{3 robots - 4 objects} &
  \multicolumn{2}{c}{4 robots - 4 objects} &
  \multicolumn{1}{c}{4 robots - 3 objects} \\ \midrule
\multicolumn{1}{l}{\textbf{Scenario}} &
  S2.a &
  \multicolumn{1}{c|}{S2.b} &
  S3.a &
  \multicolumn{1}{c|}{S3.b} &
  \multicolumn{1}{c|}{S3.c} &
  S4.a &
  \multicolumn{1}{c|}{S4.b} &
  S4.c \\ \midrule 
\multicolumn{1}{l|}{LC} &
  \textcolor{black}{0.50 $\pm$ 0.01} &
  \multicolumn{1}{c|}{{N/A}} &
  \textcolor{black}{N/A} &
  \multicolumn{1}{c|}{\textcolor{black}{N/A}} &
  \multicolumn{1}{c|}{\textcolor{black}{N/A}} &
  \textcolor{black}{N/A} &
  \multicolumn{1}{c|}{\textcolor{black}{N/A}} &
  \textcolor{black}{N/A} \\
\multicolumn{1}{l|}{TA-LC} &
  \textcolor{black}{0.50 $\pm$ 0.01} &
  \multicolumn{1}{c|}{{0.69 $\pm$ 0.20}} &
  \textcolor{black}{0.44 $\pm$ 0.03} &
  \multicolumn{1}{c|}{\textcolor{black}{0.57 $\pm$ 0.01}} &
  \multicolumn{1}{c|}{\textcolor{black}{N/A}} &
  \textcolor{black}{0.57 $\pm$ 0.01} &
  \multicolumn{1}{c|}{\textcolor{black}{N/A}} &
  \textcolor{black}{0.99 $\pm$ 0.00} \\
\multicolumn{1}{l|}{GP} &
  \textcolor{black}{0.49 $\pm$ 0.02} &
  \multicolumn{1}{c|}{\textcolor{black}{0.82 $\pm$ 0.01}} &
  \textcolor{black}{0.43 $\pm$ 0.01} &
  \multicolumn{1}{c|}{\textcolor{black}{0.53 $\pm$ 0.06}} &
  \multicolumn{1}{c|}{\textcolor{black}{N/A}} &
  \textcolor{black}{0.37 $\pm$ 0.01} &
  \multicolumn{1}{c|}{\textcolor{black}{N/A}} &
  \textcolor{black}{0.41 $\pm$ 0.01} \\
\multicolumn{1}{l|}{GP-CA} &
  \textcolor{black}{0.66 $\pm$ 0.01} &
  \multicolumn{1}{c|}{\textcolor{black}{0.82 $\pm$ 0.01}} &
  \textbf{0.45 $\pm$ 0.03} &
  \multicolumn{1}{c|}{\textcolor{black}{0.70 $\pm$ 0.03}} &
  \multicolumn{1}{c|}{\textcolor{black}{N/A}} &
  \textcolor{black}{0.39 $\pm$ 0.01} &
  \multicolumn{1}{c|}{\textcolor{black}{0.37 $\pm$ 0.02}} &
  \textcolor{black}{0.41 $\pm$ 0.01} \\
\multicolumn{1}{l|}{PuSHR} &
  \textbf{0.67 $\pm$ 0.02} & % p-value 0.035
  \multicolumn{1}{c|}{\textbf{0.94 $\pm$ 0.02}} &
  \textcolor{black}{0.42 $\pm$ 0.00} &
  \multicolumn{1}{c|}{\textbf{0.72 $\pm$ 0.01}} &
  \multicolumn{1}{c|}{\textbf{{0.73 $\pm$ 0.01}}} &
  \textbf{0.73 $\pm$ 0.01} &
  \multicolumn{1}{c|}{\textbf{{0.39 $\pm$ 0.02}}} &
  \textbf{1.07 $\pm$ 0.01} \\
\bottomrule
\end{tabular}
}
}
\end{table*}

\subsection{Results}

\begin{figure}
    \centering
    \begin{subfigure}{.49\linewidth}
        \centering
        \includegraphics[width = \linewidth]{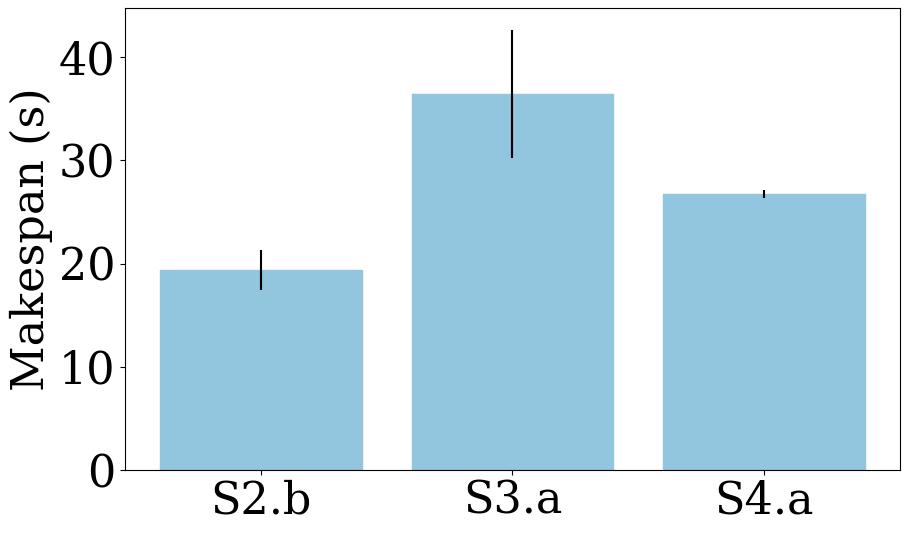}
        % \caption{\label{fig:labMakespan}}
    \end{subfigure}
    \begin{subfigure}{.49\linewidth}
        \centering
        \includegraphics[width = \linewidth]{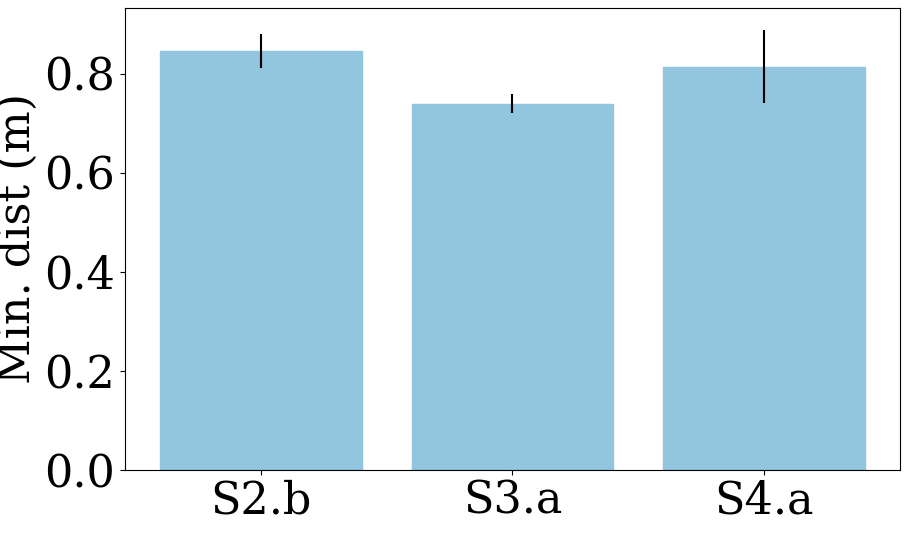}
        % \caption{\label{fig:labMinDist}}
    \end{subfigure}
    \caption{Real-world performance measured in terms of Makespan and Minimum distance. Bars and errorbars represent average values and standard deviations over 10 trials for each scenario. All trials were successful.}
  \label{fig:lab-experiments}
\end{figure}

In Tables~\ref{tab:success},~\ref{tab:makespan}, and~\ref{tab:minDist}, we report respectively the Success rate, Makespan, and Minimum distance for all algorithms across all scenarios in simulation. In~\figref{fig:lab-experiments}, we report the performance of PuSHR on three of the scenarios (S2.b, S3.a, and S4.a) executed in the lab. Footage from our lab experiments can be found at~\url{https://youtu.be/DIWmZerF_O8}.

% \textbf{H1}. GP completes 7/8 scenarios with high success (Table~\ref{tab:success}) in contrast to LC which only completes the easiest scenario (S2.a). This is because the LC strategy is reactive, lacking a sufficient horizon to anticipate conflicts in a confined space. We observed LC get trapped in dense regions, requiring tight U-turns or other complex maneuvers when $n\geq3$, eventually leading robots to exit the workspace boundary, get stuck or lose the block. %For instance, in a situation where the goal point is at the center of the robot's turning circle, the robot would keep circling the goal forever, instead of turning away first, increasing the distance between itself and the goal, and then turning back so that it has the space necessary to actually maneuver to the goal point. 
% In contrast, the inclusion of GP gives robots a better sense of the global spatial structure throughout the scenario, avoiding such artifacts.  

\textit{H1}. GP completes 7/8 scenarios with high success (Table~\ref{tab:success}) whereas LC only completes the easiest scenario (S2.a), thus H1 is confirmed. Lacking a detailed predictive model, LC cannot anticipate future conflicts in such a confined space. This observation is more pronounced for $n\geq3$ with LC often violating the workspace or getting stuck. %For instance, in a situation where the goal point is at the center of the robot's turning circle, the robot would keep circling the goal forever, instead of turning away first, increasing the distance between itself and the goal, and then turning back so that it has the space necessary to actually maneuver to the goal point. 

% \textbf{H2}. GP-CA keeps higher clearances than GP for similar success (Table~\ref{tab:minDist}), confirming H2. As $n$ increases, planning with GP produces more complex maneuvers that need to be executed in a tight space. Thus, even the slightest deviations from the plan that may naturally occur due to time indexing violations (due to computation, communication lags, etc), imperfect control tuning, etc often bring deadlocks or collisions. In real-world systems, such issues would get amplified, seriously impacting performance. We suspect that the distance gains of the GP-CA enable it to solve the challenging S4.b in contrast to GP which failed completely.

\textit{H2}. GP-CA keeps greater or equal clearances than GP for similar success (Table~\ref{tab:minDist}), thus confirming H2.
As $n$ increases, planning with GP produces more complex maneuvers that need to be executed in a tight space. Thus, minor deviations due to tracking errors can bring collisions, and CA can help avoid this by increasing clearances. The distance gains of the GP-CA enable it to solve the challenging S4.b in contrast to GP which failed completely.

% This motivated the use of a local controller focusing on collision avoidance through trajectory following and local maneuvering. Table~\ref{tab:minDist} shows the robustness of that the local collision avoidance component contributes: we see that GP-CA configuration has a slightly higher minimum distance than just the GP. 

% \begin{wrapfigure}[17]{R}{0.3\textwidth}
%   \centering
%   \vspace{-12pt}
%   \includegraphics[width=\linewidth]{figures/clcbs_makespan_vs_ecbs_soc_raw.png}
%     \caption{Task assignment quality (ECBS-TA cost) vs path quality (CL-CBS Makespan) for the $4!$ possible task assignments of scenario S4.b.\label{fig:taPathFeatureComparison}}
% \end{wrapfigure}

\begin{figure}
    \centering
    \begin{subfigure}{.48\linewidth}
        \centering
        \includegraphics[width = \linewidth]{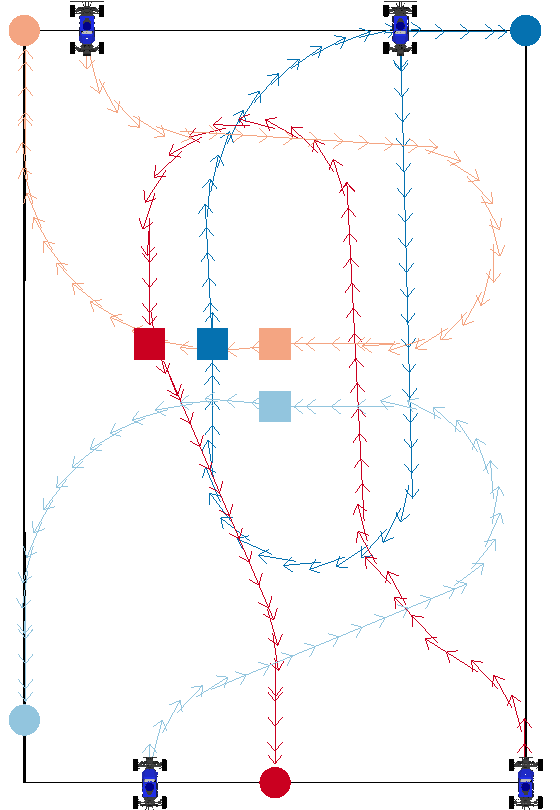}
        \caption{\label{fig:scenario4bNoTA}}
    \end{subfigure}
    \begin{subfigure}{.48\linewidth}
        \centering
        \includegraphics[width = \linewidth]{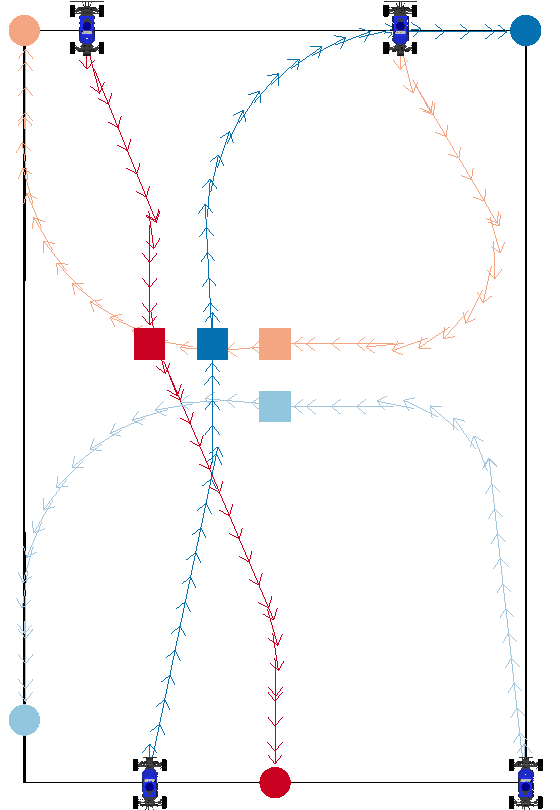}
        \caption{\label{fig:scenario4bTA}}
    \end{subfigure}
    % \begin{subfigure}{.35\linewidth}
    %     \centering
    %     \includegraphics[width = \linewidth]{figures/3c-plan-no-TA.png}
    %     \caption{\label{fig:scenario3cNoTA}}
    % \end{subfigure}
    % \begin{subfigure}{.35\linewidth}
    %     \centering
    %     \includegraphics[width = \linewidth]{figures/3c-plan-TA.png}
    %     \caption{\label{fig:scenario3cTA}}
    % \end{subfigure}
    \caption{Task simplification via task assignment (TA). (\subref{fig:scenario4bNoTA}) Paths planned by CL-CBS upon manual TA for scenario S4.b (median assignment quality out of $4!$ combinations). (\subref{fig:scenario4bTA}) Paths planned by CL-CBS upon \emph{optimal} TA from ECBS-TA for the same scenario. Blocks' initial and goal locations are noted as squares and circles of same color respectively.
    \label{fig:taSimplify}}
    % \label{fig:taSimplifyReverse}}
\end{figure}

\begin{figure}
  \centering
  \includegraphics[width=\linewidth]{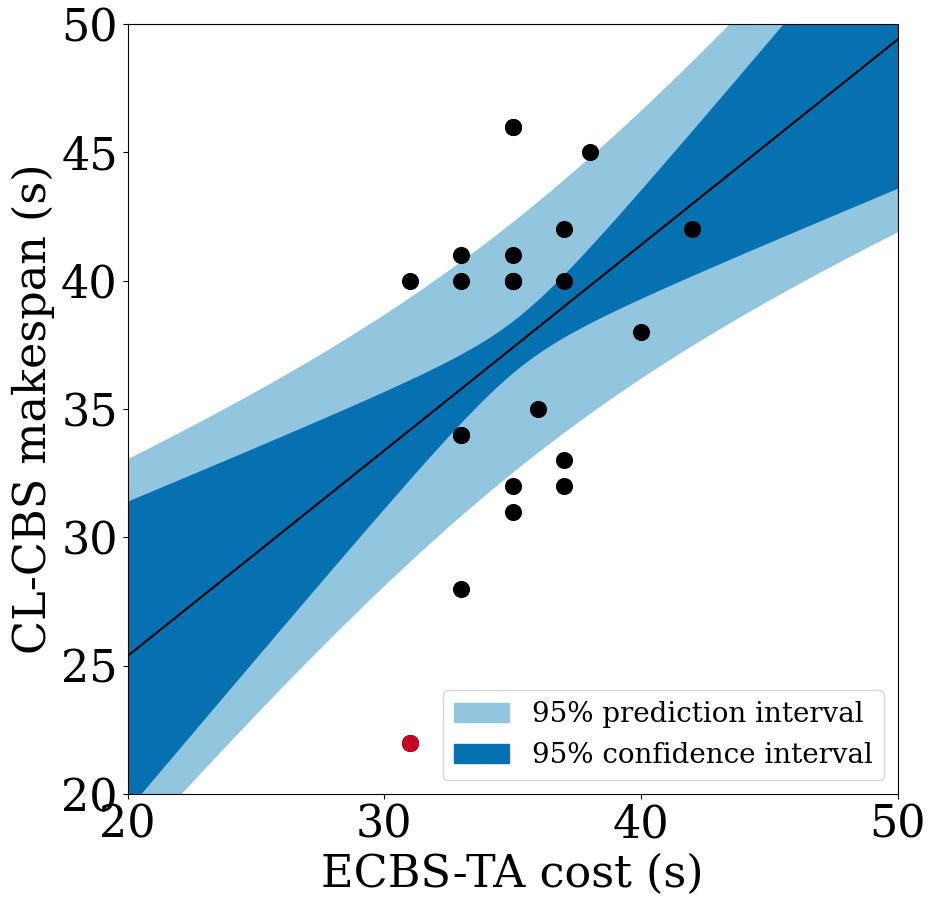}
    \caption{Task assignment quality (ECBS-TA cost) vs path quality (CL-CBS Makespan) for the $4!$ possible task assignments for scenario S4.b. PuSHR leverages optimal task assignment by ECBS-TA which enables the planner (CL-CBS) to achieve the lowest Makespan (shown in red).}
    \label{fig:taPathFeatureComparison}
\end{figure}

% In contrast, GP-CA explicitly accounts for such imperfect trajectory tracking by correcting for imminent collisions. 
% I feel like this is hypothesis overloading but oh well, what can you do?
\textit{H3}. PuSHR and TA-LC improve respectively over GP-CA and LC (Table~\ref{tab:success}) confirming H3. We also see that PuSHR and TA-LC achieve improved Makespans (Table~\ref{tab:makespan}) suggesting that TA enables more efficient coordination among agents. Taking a deeper look, we see that this is indeed the case: \figref{fig:taPathFeatureComparison} shows a positive correlation between TA quality (ECBS-TA cost) and path planning quality (CL-CBS Makespan), with the lowest ECBS-TA cost coinciding with the lowest CL-CBS Makespan. This pattern is also visible qualitatively in~\figref{fig:taSimplify}: plans generated upon TA are geometrically simpler. This simplification is helpful in more complex cases such as S3.c, where the non-TA paths are too difficult for the MPC to track properly, causing even GP-CA to fail.

\textit{H4}. Thanks to TA and collision avoidance, PuSHR scales best as the only one to succeed in all scenarios, always with the top success rate. This confirms H4. Our system is slower than TA-LC in a few scenarios (S3.a, S3.b, S4.a, s4.c) but more successful as the global plan often pauses the trajectories to synchronize robots whereas TA-LC does not, thus finishing faster in simple scenarios. It is important to note that these scenarios are much simpler than e.g., S4.b, involving sparse regions, and do not require much turning (see~\figref{fig:scenarios}). 

% TA-LC also takes more direct paths since it doesn't consider block approach angle, so its path is shorter than with a global plan which includes turns/curves for contact at the right angle. 

% We see that although our system is slower than TA-LC and less distant in a few scenarios, it is the only system to complete all scenarios, and it does so with a success rate $\geq 99\%$. Thus, H4 is confirmed.

% Overall, our system exhibited robust performance that scaled well with the number of robots across both simulated and real-world experiments. 

\textit{H5}. Our system achieved 100\% success in the lab trials (\figref{fig:lab-experiments}), which implies that the robots were consistently able to maintain control of their blocks and push them to their goals. Thus H5 is confirmed. Compared to the simulated trials, we see that in the lab the Makespan increases, an artifact of communication, computation, and actuation overhead. The Minimum distance varies but provides a sufficient buffer to avoid collisions. We observed that the collision avoidance feature of the MPC was instrumental in making sure the robots succeeded even when unmodeled physics or communication delays caused deviations from the planned trajectories.

\section{Discussion}

\textbf{Task assignment}. Our findings (H3) confirmed the value of task assignment for our domain. We suspected this when we saw that a manual assignment strategy using a distance-based heuristic for task assignment would fail when distances are roughly equal (e.g.,~\ref{fig:scenario4b}). This motivated us to incorporate ECBS-TA \citep{ecbsta} into our system which significantly boosted performance. However, the ECBS-TA implementation makes use of a Euclidian-distance-based heuristic, which is unaware of robots' kinematics or pushing constraints and does not guarantee ideal assignments as the workspace density increases. A direction for future work is to design a heuristic that efficiently incorporates such constraints.

% This will likely result in non-advantageous assignments as the complexity of the workspace increases. The consideration of smooth curve functions like splines or Dubins curves could help provide more relevant distance costs to guide the task assignment. 

\textbf{Tractability}. As we initially experimented with ECBS-TA and CL-CBS, two CBS-driven algorithms, we explored the idea of jointly extracting task assignments and trajectories through CL-CBS. However, simultaneously iterating over multiagent trajectories and assignments using the CBS paradigm is intractable. In contrast, ECBS-TA finds the optimal task assignment very efficiently by planning in a much lower resolution (a 2-dimensional grid); then CL-CBS produces a single joint plan in the higher-resolution space that leverages the simplification induced by the advantageous assignment. For reference, planning for the S4.a scenario (\figref{fig:scenarios}) takes around $10min$ with the former approach whereas PuSHR takes about 10 seconds. Computations took place on a laptop with an Intel i7-10750H CPU (6 cores @ 2.6 GHz). This enabled us to scale our evaluation and run hardware experiments. However, the problem of efficiently combining task and motion planning in this domain is still interesting and worth investigating.

%\Our approach uses two conflict based search(CBS) approaches, first at the task assignment level and then at the global planning level. The reason for not combining both into a single CBS system is the computational cost of performing optimal task assignment with car-like kinematic constraints. Note that our task assignment system does in-fact use the actual planning costs to find the optimal task assignment, as done by~\citet{ecbsta}, but it plans on a low resolution 2-dimensional grid space. The resulting task assignment is then used by the global planner to produce higher resolution paths while considering the car-like kinodynamics. As this step takes a fair amount of time (6-8 seconds on a modern desktop), and task assignment with actual path costs would have to repeatedly solve the MAPF problem for different task assignments, the proposed architecture begins to make sense.

%\cmnote{One could ask: Our approach essentially invokes 2 different CBS-based approaches for path planning --first to assign tasks, 2nd to get paths. Why are we not invoking CBS once to get both assignments and paths? In other words, is there an easy way to modify ECBS-TA OR CL-CBS to do both task assignment AND path planning? If not why? How do we justify this "planning duplicacy"}

\textbf{Quasistatics}. Leveraging stable pushing~\citep{lynch1996}, PuSHR completed challenging tasks under constraints imposed by robots' kinematics, their plans, and the workspace boundary without closing the loop for block control. While more finegrained contact models~\citep{pushnet,Bauza2017,Stber2020LetsPT} could assist in scaling towards more extreme settings (e.g., dynamic pushing, irregular objects), our experiments demonstrate the value of quasistatic analytical modeling for a variety of practical problems involving contact. %We further note that an additional cost in the MPC for maintaining the block at the center of the bumper could allow the system to operate outside the quasistatic regime.
%\atnote{Mention additional feedback cost for block location when leaving quasistatic regime as possible future work?}

% \atnote{talk about 0.5m align distance}

% \atnote{talk about MPC multiagent collision cost (vs no collision cost)}

% \cmnote{talk about how we remove the turning radius constraint before contact}

% \subsection{Limitations}

% Further, consideration of the nonholonomic constraints would also allow the task assignment system to provide the optimal orientation at the pickup/delivery locations, something that currently requires human intervention.

% \cmnote{implementation of manual assignment as a limitation}

\balance
\bibliography{references}
\bibliographystyle{abbrvnat}

% \appendix
% \input{sections/appendix.tex}

\end{document}